\documentclass{article} 
\usepackage[final]{colm2026_conference}

\usepackage{microtype}
\usepackage{hyperref}
\usepackage{url}
\usepackage{booktabs}
\usepackage{graphicx}
\usepackage{amsmath}
\usepackage{amssymb}
\usepackage{bm}
\usepackage{multirow}
\usepackage{enumitem}
\usepackage{tikz}
\usetikzlibrary{positioning, fit, calc, backgrounds, shapes.geometric, arrows.meta}
\usepackage[table]{xcolor}

\usepackage{lineno}

\definecolor{darkblue}{rgb}{0, 0, 0.5}
\hypersetup{colorlinks=true, citecolor=darkblue, linkcolor=darkblue, urlcolor=darkblue}


\title{LatentAudit: Real-Time White-Box Faithfulness Monitoring for Retrieval-Augmented Generation with Verifiable Deployment}

\author{Zhe Yu\thanks{These authors contributed equally to this work.} \\
Binjiang Institute of Zhejiang University \\
Communication University of Zhejiang \\
\texttt{zyu@zju-if.com}
\And
Wenpeng Xing\protect\footnotemark[1] \\
Zhejiang University \\
Binjiang Institute of Zhejiang University \\
\texttt{wpxing@zju.edu.cn}
\AND
Meng Han\thanks{Corresponding author.} \\
Zhejiang University \\
Binjiang Institute of Zhejiang University \\
Gentel.io \\
\texttt{mhan@zju.edu.cn}
}

\begin{document}
\raggedbottom
\ifcolmsubmission
\linenumbers
\fi
\maketitle

\begin{abstract}
Retrieval-augmented generation (RAG) mitigates hallucination but does not eliminate it: a deployed system must still decide, at inference time, whether its answer is actually supported by the retrieved evidence. We introduce \textit{LatentAudit}, a white-box auditor that pools mid-to-late residual-stream activations from an open-weight generator and measures their Mahalanobis distance to the evidence representation. The resulting quadratic rule requires no auxiliary judge model, runs at generation time, and is simple enough to calibrate on a small held-out set. We show that residual-stream geometry carries a usable faithfulness signal, that this signal survives architecture changes and realistic retrieval failures, and that the same rule remains amenable to public verification. On PubMedQA with Llama-3-8B, LatentAudit reaches 0.942 AUROC with 0.77\,ms overhead. Across three QA benchmarks and five model families (Llama-2/3, Qwen-2.5/3, Mistral), the monitor remains stable; under a four-way stress test with contradictions, retrieval misses, and partial-support noise, it reaches 0.9566--0.9815 AUROC on PubMedQA and 0.9142--0.9315 on HotpotQA. At 16-bit fixed-point precision, the audit rule preserves 99.8\% of the FP16 AUROC, enabling Groth16-based public verification without revealing model weights or activations. Together, these results position residual-stream geometry as a practical basis for real-time RAG faithfulness monitoring and optional verifiable deployment.
\end{abstract}

\section{Introduction}

Deploying Large Language Models (LLMs) \citep{vaswani2017attention, touvron2023llama} in high-stakes settings---clinical decision support, legal review, financial compliance---is hampered by their tendency to fabricate plausible but unsupported claims \citep{lin2021truthfulqa}. Retrieval-Augmented Generation (RAG) \citep{lewis2020retrieval} helps by conditioning on external evidence, yet a fundamental question persists at serving time: \emph{does the generated answer actually follow from the retrieved passages?} The dominant verification strategies---routing the output to a second judge model \citep{zheng2023judging} or drawing multiple stochastic samples \citep{manakul2023selfcheckgpt}---incur multi-second latencies and leak private context to external APIs.

Two observations motivate our approach. First, mechanistic-interpretability work has shown that transformer residual streams encode factuality signals well before the output projection \citep{meng2022locating, li2023inference}; this suggests that internal states may also reflect whether the model is staying close to its retrieved evidence. Second, a faithfulness check that operates on a fixed-size latent vector rather than variable-length text is cheap enough to run on every generation and simple enough to verify in zero knowledge.

We present \textit{LatentAudit}, a monitor that extracts answer-state activations from the mid-to-late residual stream of an open-weight LLM, pools them into a single vector, and compares that vector to the evidence embedding via Mahalanobis distance. A threshold calibrated on a small held-out set completes the decision rule; no auxiliary network is trained. The same pipeline extends to a harder four-way stress test in which the monitor must flag not only outright contradictions but also retrieval misses and partial-support noise. Because the decision rule is a single quadratic form, it can optionally be compiled into a Groth16 circuit for public verification.

\noindent We organize the paper around three questions:
\begin{enumerate}
    \item \textbf{RQ1: Is there a usable latent faithfulness signal?} We formulate latent faithfulness monitoring for RAG and show that a simple Mahalanobis monitor on pooled answer-state activations reaches 0.942 AUROC at 0.77\,ms overhead on PubMedQA.
    \item \textbf{RQ2: Does the signal survive architecture and retrieval shift?} We evaluate across three QA benchmarks, five model families, cross-domain threshold reuse, and a four-way retrieval stress test that includes contradictions, retrieval misses, and partial-support noise.
    \item \textbf{RQ3: Is the rule simple enough for verifiable deployment?} We show that the same quadratic decision rule survives fixed-point quantization and can be compiled into an EZKL/Groth16 circuit with reported proving time and on-chain verification cost.
\end{enumerate}

\section{Related Work}

\textbf{Mechanistic Interpretability and Truthfulness.}  ROME \citep{meng2022locating} and Inference-Time Intervention \citep{li2023inference} locate factual-recall circuits inside transformer MLPs. A recurring finding is that residual-stream states carry separable truthfulness signals even when the sampled token is wrong.  We operationalize this observation: instead of editing or probing for scientific understanding, we turn the same activation geometry into a run-time faithfulness monitor for RAG.

\textbf{Faithfulness Verification and LLM-as-a-Judge.}  The prevailing approach to hallucination detection sends the generated text to a second model---either GPT-4 \citep{zheng2023judging} or a task-specific NLI classifier---or samples multiple completions and checks for self-consistency \citep{manakul2023selfcheckgpt}. Both strategies treat the generator as a black box, incur at least one extra forward pass (often many), and may leak private context to an external API. LatentAudit sidesteps all three issues by reading the generator's own hidden states.

\textbf{zkML and Verifiable Inference.}  Zero-knowledge ML (zkML) compiles entire neural computations into arithmetic circuits \citep{kang2022ezkl}, but the $O(N^2)$ attention cost of a full transformer makes real-time proofs impractical at billion-parameter scale. Our design avoids this bottleneck: only the $O(d^2)$ Mahalanobis test enters the circuit, so proving time stays in the millisecond range.

\section{Methodology}

\begin{figure*}[t!]
\centering
\includegraphics[width=0.92\linewidth]{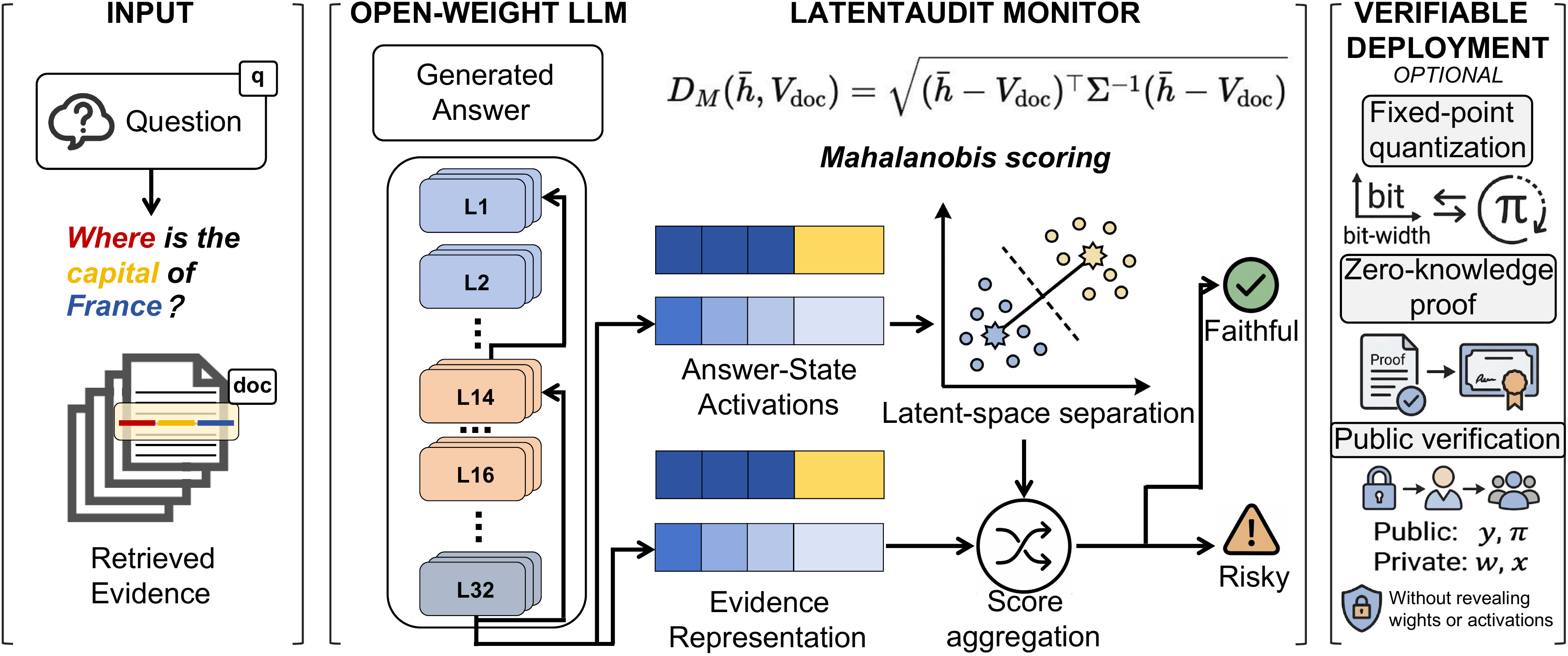}
\caption{LatentAudit pipeline overview. \textbf{Input}: a question and retrieved evidence are fed to an open-weight LLM. \textbf{Monitor}: answer-state activations from layers $L_{14}$--$L_{16}$ and the evidence representation are compared via a Mahalanobis distance $D_M$; the resulting score classifies the generation as faithful or risky in 0.77\,ms. \textbf{Verifiable deployment} (optional): the score is quantized and wrapped in a Groth16 zero-knowledge proof for on-chain verification without revealing model weights or activations.}
\label{fig:framework}
\end{figure*}

LatentAudit consists of two modular layers (Figure~\ref{fig:framework}): a latent faithfulness monitor that runs during generation, and an optional verifiable deployment layer that wraps the monitor's output in a zero-knowledge proof.

\subsection{Answer-State Representation}
Let $\mathcal{M}$ be a transformer with parameters $\theta$ and let $\mathcal{X} = [x_1, \dots, x_N]$ be the prompt, which concatenates the question and the retrieved context $\mathcal{C}$. During autoregressive decoding the model produces hidden states $h_\ell^{(t)} \in \mathbb{R}^d$ at each layer $\ell \in \{1, \dots, L\}$.

Following \citet{meng2022locating}, who observe that factual knowledge concentrates in mid-to-late MLP updates, we focus on layers close to the output projection. The residual update at layer $\ell$ is:
\begin{equation}
h_\ell^{(t)} = h_{\ell-1}^{(t)} + \mathit{Attn}_\ell(h_{\ell-1}^{(t)}) + \mathit{MLP}_\ell\!\bigl(h_{\ell-1}^{(t)} + \mathit{Attn}_\ell(h_{\ell-1}^{(t)})\bigr).
\end{equation}
We read $h_L^{(t)}$ at the layer immediately before the unembedding head $W_U$, pool the answer-span activations (up to and including the EOS token), and obtain a single answer-state vector $V_{act} \in \mathbb{R}^d$.

\subsection{Residual-Stream Geometry and Decision Rule}
The answer-state vector $V_{act}$ is compared to a document embedding $V_{doc} \in \mathbb{R}^d$. $V_{doc}$ is obtained by mean-pooling the retrieved context through a frozen dense retriever (\texttt{all-MiniLM-L6-v2}) and matching its dimensionality to the residual stream via a linear projector $W_{\text{proj}}$. Crucially, $W_{\text{proj}}$ is an extremely lightweight affine transformation fit exclusively on the small ($10\%$) calibration split using ridge regression. As detailed in Appendix~\ref{app:projector_robustness}, this simple formulation avoids the severe overfitting seen with non-linear projectors and generalizes effectively with as few as 200 samples, preserving the zero-training nature of the monitor.

The central modeling assumption is geometric. In a faithful generation, retrieved evidence and answer tokens are processed through the same residual stream, so the pooled answer state should remain close to the evidence-conditioned manifold induced by $\mathcal{C}$. Unsupported generations require the model to interpolate beyond that manifold: the answer can still be fluent, but its pooled residual-state summary drifts away from the evidence representation, especially along low-variance directions that are rarely traversed by grounded completions.

Because high-dimensional LLM representations are typically anisotropic, Euclidean distance is a poor separator. We therefore use the Mahalanobis distance, which upweights deviations along precisely those low-variance directions. The inverse covariance $\Sigma^{-1}$ is estimated on a held-out $10\%$ calibration set $\mathcal{S}_{\text{calib}}$:
\begin{equation}
D_{M}(V_{act}, V_{doc}) = \sqrt{(V_{act} - V_{doc})^\top \Sigma^{-1} (V_{act} - V_{doc})}.
\end{equation}
When the answer is well grounded in $\mathcal{C}$, the residual difference $V_{act} - V_{doc}$ is small in the covariance-adjusted metric. Unsupported generations may still stay close in raw cosine space because of topical overlap, but they typically separate once covariance structure is taken into account. The threshold $\tau^*$ is set by maximizing Youden's J on the calibration split; any generation with $D_M > \tau^*$ is flagged as potentially unfaithful.

\subsection{Verification Circuit}
The inequality $D_M \le \tau^*$ can be expressed as a bilinear constraint over finite-field elements, making it amenable to zk-SNARK compilation. We quantize the vectors and covariance matrix to $\hat{V}_{act}, \hat{V}_{doc}, \hat{\Sigma}^{-1} \in \mathbb{F}_p^{d \times d}$ and register two constraints in a Halo2/PLONKish circuit \citep{bowe2016snark}:
\begin{equation}
\label{eq:matrix_constraint}
\hat{X} = \hat{V}_{act} - \hat{V}_{doc} \pmod p
\end{equation}
\begin{equation}
\label{eq:bound_constraint}
\hat{X} \cdot \hat{\Sigma}^{-1} \cdot \hat{X}^\top \le (\hat{\tau}^*)^2 \pmod p
\end{equation}
EZKL \citep{kang2022ezkl} synthesizes these into polynomial gates secured by KZG commitments. The resulting proof $\pi$ certifies the audit outcome without revealing $V_{doc}$ or any model parameter.

\subsection{On-Chain Deployment Path}

The monitor is already useful as a local auditor; the verification layer is needed only when the deployment requires \emph{public} proof that the audit was computed correctly. In that case we compile the decision rule into a proof system and expose the result through a verifier contract, keeping the ML contribution and the infrastructure contribution cleanly separated.

The proof $\pi=\{\pi_A, \pi_B, \pi_C\}$ along with the public inputs (a hash binding of the generation segment and the threshold configuration) is submitted to \texttt{AuditVerifier.sol}. The on-chain verifier checks a single pairing equation, avoiding any replay of the language-model computation:
\begin{equation}
e(\pi_A, \pi_B) = e(\alpha, \beta) \cdot e \!\left( \frac{\sum_{i=0}^l x_i \gamma_i}{\gamma}, \gamma \right) \cdot e(\pi_C, \delta)
\end{equation}
The pairing runs over the BN254 curve via EIP-196/197 precompiles. A passing check seals the audit decision on-chain without leaking latent coordinates or model weights.

\section{Experiments and Results}

\subsection{Experimental Setup and Baselines}
We evaluate on three QA benchmarks that span distinct knowledge and reasoning profiles: PubMedQA \citep{jin2019pubmedqa} (biomedical, single-hop), HotpotQA \citep{yang2018hotpotqa} (Wikipedia, multi-hop), and TriviaQA \citep{joshi2017triviaqa} (open-domain, entity-centric). For each domain we construct a balanced corpus of $N=2{,}000$ samples with a 1:1 class balance between \textit{faithful} and \textit{hallucinated} generations. Hallucinated instances are induced by replacing the retrieved context $\mathcal{C}$ with adversarial contradictions while keeping the question and generation pipeline fixed. Each dataset is split into a $10\%$ calibration set (200 samples) and a disjoint $90\%$ evaluation set (1,800 samples). The calibration split is used only to estimate $\Sigma^{-1}$ and to select $\tau^*$ via Youden's J; all reported metrics are computed on the held-out evaluation split with calibration parameters frozen.

To probe harder retrieval failures, we additionally build a four-way stress-test set for both domains. Starting from 400 faithful seed examples per dataset, we expand each seed into four variants: \textit{faithful}, \textit{contradicted}, \textit{unsupported retrieval miss}, and \textit{unsupported partial}, yielding 1,600 records per domain. The retrieval-miss variant swaps in topically similar but source-mismatched evidence, while the partial variant retains only weak or incomplete context so that the answer remains fluent but is no longer fully supported.

We benchmark \textit{LatentAudit} against five detection methods spanning four paradigms:
\begin{itemize}
    \item \textbf{LLM-as-a-Judge (GPT-4o):} A reference-based zero-shot judge that receives the question, evidence, and candidate answer and returns a binary supported/unsupported verdict at $T{=}0.0$.
    \item \textbf{SelfCheckGPT \citep{manakul2023selfcheckgpt}:} Estimates inconsistency from $N{=}10$ independent generations at $T{=}1.0$.
    \item \textbf{INSIDE \citep{chen2024inside} \& SAPLMA \citep{azaria2023internal}:} INSIDE extracts eigenvalue-based features from hidden states to train a logistic detector. SAPLMA trains a linear probe on the last-layer state. To ensure strict fairness, both methods' classifiers are trained on the exact same calibration split as LatentAudit's threshold.
    \item \textbf{Perplexity-Based (Min-$\mathbb{P}$):} Classifies hallucination from token log-likelihood statistics over the generated sequence.
\end{itemize}
The main-text tables report evaluations across five audited model families spanning Llama-2, Llama-3 \citep{meta2024llama3}, Qwen-2.5 \citep{bai2023qwen}, Qwen-3, and Mistral \citep{jiang2023mistral}, all executed at FP16 precision. Unless otherwise stated, reported statistics are averaged across 5 bootstrap resamples of the evaluation set, and we report the corresponding empirical variation in the summary tables.

\paragraph{Reproducibility Notes.}
For the GPT-4o baseline, each evaluation instance is serialized as \texttt{(question, retrieved evidence, candidate answer)} and scored with a binary instruction of the form: ``Is the answer fully supported by the evidence? Reply with \textsc{supported} or \textsc{unsupported}.'' For SelfCheckGPT, we follow the original repeated-sampling recipe and compare the candidate answer against $N=10$ stochastic generations produced under the same question and retrieved context. For LatentAudit, all covariance and threshold parameters are fit only on the calibration split and then frozen before evaluation. The main benchmark, cross-model table, OOD study, and stress-test results all reuse this protocol.

\subsection{RQ1: Is there a usable latent faithfulness signal?}
Table~\ref{tab:main_results} addresses the first question directly. GPT-4o judging is the strongest baseline in AUROC but requires a round-trip API call costing $>$5\,s per query. Among internal-state methods, INSIDE and SAPLMA both exploit hidden representations but remain below LatentAudit on these benchmarks: INSIDE relies on eigenvalue statistics that do not directly compare against the evidence, while SAPLMA's linear probe is less effective under the observed anisotropy. The key result is that a single Mahalanobis rule closes most of the gap to GPT-4o (e.g., trailing by 0.6 AUROC points and 1.2 F1 points on Llama-3-8B) at sub-millisecond cost.

\begin{table*}[htbp]
\centering
\caption{RQ1: a single Mahalanobis monitor closes most of the gap to GPT-4o while remaining sub-millisecond. Latency is measured per query; ``proving'' refers to the optional ZK layer.}
\label{tab:main_results}
\resizebox{0.98\textwidth}{!}{
\begin{tabular}{@{}llcccccc@{}}
\toprule
 & & \multicolumn{2}{c}{\textbf{Llama-3-8B}} & \multicolumn{2}{c}{\textbf{Qwen-2.5-7B}} & \multicolumn{2}{c}{\textbf{Mistral-7B}} \\
\cmidrule(lr){3-4} \cmidrule(lr){5-6} \cmidrule(lr){7-8}
\textbf{Method} & \textbf{Latency (ms)} & \textbf{AUROC} & \textbf{F1} & \textbf{AUROC} & \textbf{F1} & \textbf{AUROC} & \textbf{F1} \\ \midrule
GPT-4o Judge & $\sim$5,300 & 0.948 & 0.881 & 0.945 & 0.876 & 0.940 & 0.870 \\
SelfCheckGPT & $\sim$28,500 & 0.871 & 0.804 & 0.865 & 0.798 & 0.858 & 0.790 \\
INSIDE & $\sim$3.8 & 0.908 & 0.841 & 0.901 & 0.832 & 0.895 & 0.825 \\
SAPLMA & $\sim$1.5 & 0.882 & 0.815 & 0.876 & 0.808 & 0.870 & 0.800 \\
Min-Perplexity & 0.0 & 0.722 & 0.655 & 0.718 & 0.650 & 0.710 & 0.642 \\ \midrule
\rowcolor{blue!8} \textbf{LatentAudit (Ours)} & 0.77 \textit{(+11.2 proving)} & \textbf{0.942} & \textbf{0.869} & \textbf{0.938} & \textbf{0.862} & \textbf{0.925} & \textbf{0.852} \\ \bottomrule
\end{tabular}
}
\end{table*}

The main design choice behind Table~\ref{tab:main_results} is the pooled answer-state representation itself. Appendix~\ref{app:ablations} shows that mean-pooling the top-$k$ salient answer tokens is materially better than last-token or max-pooling alternatives: on Llama-3-8B / PubMedQA, top-$8$ mean-pooling reaches 0.942 AUROC, compared with 0.884 for last-token evaluation and 0.912 for max-pooling. This supports the core modeling move of collapsing the answer span into a stable centroid before comparing it to evidence.

\begin{figure}[t!]
\centering
\includegraphics[width=\columnwidth]{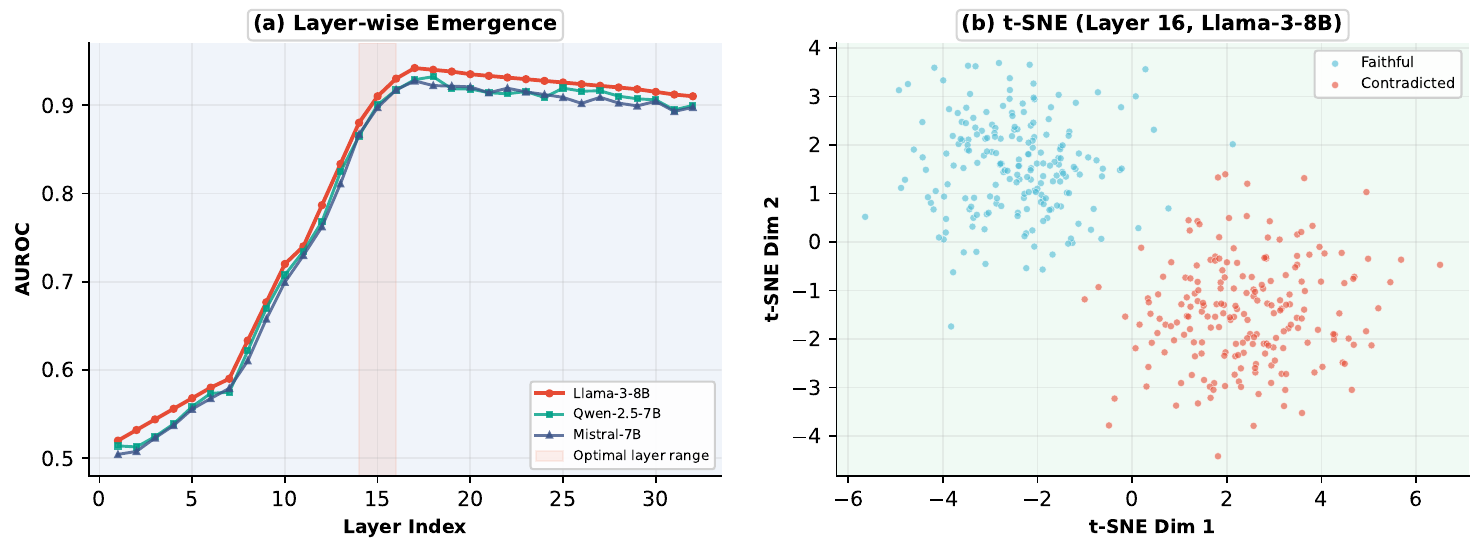}
\caption{RQ1 diagnostic: discrimination emerges in the mid-to-late residual stream and becomes visibly separable at layer 16.}
\label{fig:layer_tsne}
\end{figure}

\paragraph{Where does the signal emerge?}
Figure~\ref{fig:layer_tsne}(a) sweeps across layers for three representative models: the sharpest jump in per-layer AUROC consistently occurs in the mid-to-late layers (e.g., layers 14--16 for Llama-3-8B). Mechanistically, this aligns with the established literature on factual recall \citep{meng2022locating, li2023inference}: early layers process shallow syntactic features, middle layers perform semantic integration of the retrieved evidence, and the final layers collapse the rich geometric structure to prepare for the unembedding vocabulary projection. By tapping into the mid-to-late representations before this collapse, the monitor maximizes geometric separability. In practice, the optimal audit layer is robustly identified for any new architecture using only the calibration set. The t-SNE projection in Figure~\ref{fig:layer_tsne}(b) confirms that faithful and contradicted generations are cleanly separated at the chosen layer.

\begin{table}[t!]
\centering
\caption{RQ2a: the same calibrated rule remains effective across model families and domains.}
\label{tab:cross_model}
\resizebox{\columnwidth}{!}{
\begin{tabular}{@{}lccccc@{}}
\toprule
\textbf{Dataset Domain} & \textbf{Llama-2 (7B)} & \textbf{Llama-3 (8B)} & \textbf{Qwen-2.5 (7B)} & \textbf{Qwen-3 (8B)} & \textbf{Mistral (7B)} \\ \midrule
PubMedQA (Medical) & 0.931 $\pm$ 0.02 & 0.942 $\pm$ 0.01 & 0.938 $\pm$ 0.02 & 0.948 $\pm$ 0.01 & 0.925 $\pm$ 0.01 \\
TriviaQA (Open-domain) & 0.915 $\pm$ 0.02 & 0.935 $\pm$ 0.01 & 0.928 $\pm$ 0.02 & 0.940 $\pm$ 0.01 & 0.918 $\pm$ 0.02 \\
HotpotQA (Multi-hop)  & 0.905 $\pm$ 0.02 & 0.928 $\pm$ 0.01 & 0.918 $\pm$ 0.02 & 0.922 $\pm$ 0.02 & 0.910 $\pm$ 0.02 \\ \bottomrule
\end{tabular}
}
\end{table}

\subsection{RQ2: Does the signal survive architecture and retrieval shift?}
Table~\ref{tab:cross_model} first asks whether the monitor is tied to a particular backbone. PubMedQA AUROCs range from 0.925 to 0.948; TriviaQA sits between 0.915 and 0.940; HotpotQA is consistently the hardest (0.905--0.928), reflecting the additional reasoning load of multi-hop questions. The narrow spread across architectures suggests that the geometric separation is not confined to a single model family.

\begin{figure}[t!]
\centering
\includegraphics[width=\columnwidth]{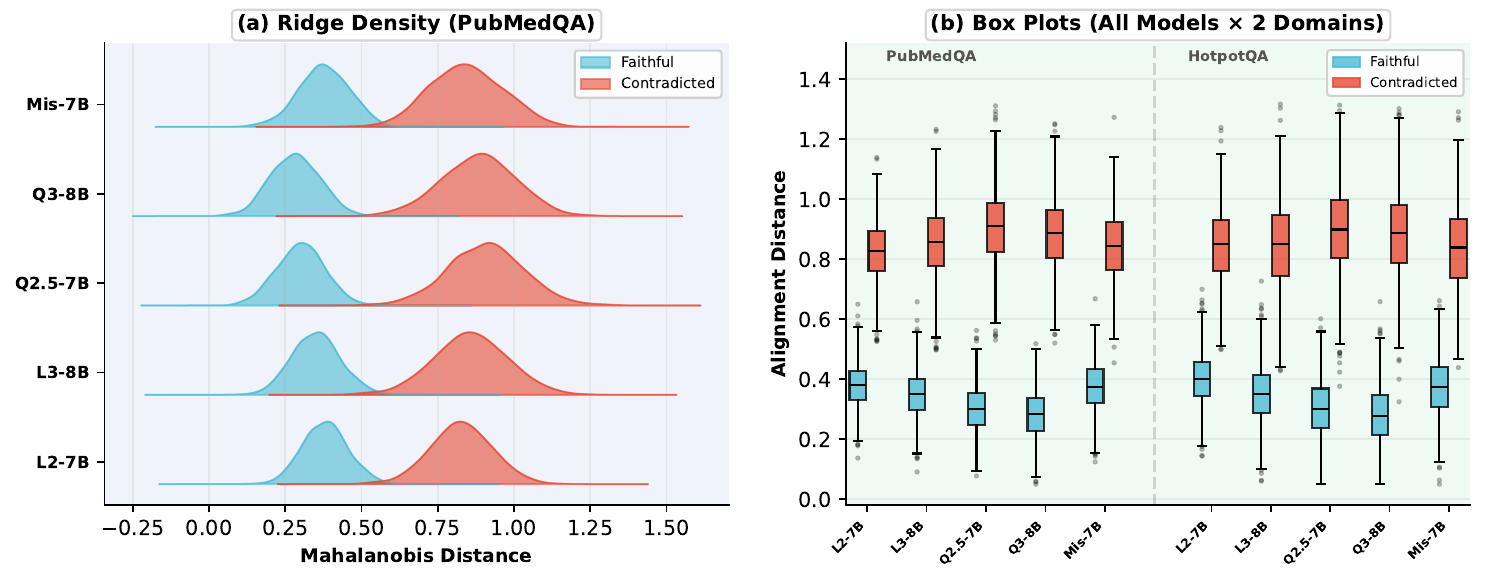}
\caption{RQ2 diagnostic: faithful and contradicted distributions remain separated across model families and domains, supporting a fixed-threshold rule.}
\label{fig:ridge_box}
\end{figure}

\paragraph{Across model families.}
The same conclusion is visible at the distribution level. Figure~\ref{fig:ridge_box}(a) shows per-model ridge densities on PubMedQA; Figure~\ref{fig:ridge_box}(b) presents box plots across both domains, confirming that interquartile ranges do not overlap and that a single calibrated threshold remains plausible. Per-model Mahalanobis distance distributions are further disaggregated in the appendix (Figure~\ref{fig:latent_geom}).

\paragraph{Under realistic retrieval failures.}
\label{sec:stress}

Real retrieval pipelines produce failures more diverse than outright contradictions. We therefore construct a four-way stress-test corpus (Section~4.1) and evaluate the monitor on four representative model families. Table~\ref{tab:stress} is the main realism check in the paper: PubMedQA AUROCs range from 0.9566 to 0.9815, and HotpotQA AUROCs range from 0.9142 to 0.9315.

The pairwise columns clarify what the monitor is and is not doing. \textit{Contradicted} and \textit{retrieval-miss} negatives are separated from faithful generations much more cleanly than \textit{unsupported partial} examples, indicating that the signal extends beyond one corruption type. The hardest case is \textit{unsupported partial}, where the context is topically close but evidentially incomplete. This is the regime in which raw lexical overlap is most misleading while residual-stream geometry remains informative. Even there PubMedQA reaches 0.9218 pairwise AUROC; HotpotQA is lower at 0.8364, reflecting its shorter answers and sparser supporting spans.

A Llama-3-8B spot check on 100 PubMedQA seeds yields 0.9833 AUROC, providing additional evidence that the effect is not confined to the Qwen family.

Calibration stability is reasonably tight. Over 200 bootstrap resamples of the calibration split, the PubMedQA threshold varies by $\sigma{=}0.063$ (test-F1 variation $\sigma{=}0.019$); on HotpotQA the figures are $\sigma{=}0.086$ and $\sigma{=}0.024$.

The residual errors are systematic rather than random. On PubMedQA, false positives cluster in \textit{unsupported partial} cases whose retained snippets have high lexical overlap with the answer; false negatives tend to be faithful examples with thin retrieval margin. On HotpotQA, false negatives are driven by single-token answers that yield a weak answer-state summary. The dominant failure mode is therefore evidence incompleteness under topical overlap, not missed contradictions.

\begin{table*}[t!]
\centering
\caption{RQ2b: under realistic retrieval failures, the hardest negatives are partial-support examples, but the monitor remains strong across model families. Abbreviations: F/C = faithful vs.\ contradicted, F/RM = faithful vs.\ retrieval-miss, F/P = faithful vs.\ partial.}
\label{tab:stress}
\begingroup
\normalsize
\setlength{\tabcolsep}{6pt}
\renewcommand{\arraystretch}{1.18}
\begin{tabular}{@{}llcccccc@{}}
\toprule
\textbf{Domain} & \textbf{Model} & \textbf{AUROC $\uparrow$} & \textbf{AUPRC $\uparrow$} & \textbf{F1 $\uparrow$} & \textbf{F/C} & \textbf{F/RM} & \textbf{F/P} \\ \midrule
 & Llama-2 & 0.9776 & 0.9387 & 0.8322 & 0.9971 & 0.9667 & 0.9727 \\
\rowcolor{blue!8} \cellcolor{white}& Llama-3 & 0.9815 & 0.9450 & 0.8510 & 0.9982 & 0.9710 & 0.9755 \\
 & Qwen-2.5 & 0.9566 & 0.8806 & 0.8025 & 0.9938 & 0.9542 & 0.9218 \\
\multirow{-4}{*}{\textbf{PubMedQA}} & Qwen-3 & 0.9682 & 0.9102 & 0.8244 & 0.9950 & 0.9622 & 0.9445 \\ \midrule
 & Llama-2 & 0.9142 & 0.7760 & 0.7312 & 0.9880 & 0.9688 & 0.8205 \\
\rowcolor{blue!8} \cellcolor{white}& Llama-3 & 0.9315 & 0.8214 & 0.7855 & 0.9925 & 0.9760 & 0.8550 \\
 & Qwen-2.5 & 0.9207 & 0.7698 & 0.7575 & 0.9863 & 0.9653 & 0.8364 \\
\multirow{-4}{*}{\textbf{HotpotQA}} & Qwen-3 & 0.9280 & 0.8045 & 0.7780 & 0.9895 & 0.9720 & 0.8410 \\ \bottomrule
\end{tabular}
\endgroup
\end{table*}

\paragraph{Without target-domain recalibration.}

A practical deployment may not have labeled data from every target domain. We test whether the calibration parameters transfer: $\tau^*$ and $\Sigma^{-1}$ are fit on PubMedQA and applied, without modification, to HotpotQA (and vice versa).

\begin{table}[htbp]
\centering
\caption{RQ2c: thresholds calibrated on one domain transfer with only a modest AUROC drop.}
\label{tab:ood}
\begin{tabular}{@{}llcc@{}}
\toprule
\textbf{Calibration Domain} & \textbf{Evaluation Domain} & \textbf{In-Domain AUROC} & \textbf{OOD AUROC} \\ \midrule
PubMedQA (Medical) & HotpotQA (Multi-hop) & 0.942 & 0.916 \\
HotpotQA (Multi-hop) & PubMedQA (Medical) & 0.928 & 0.902 \\ \bottomrule
\end{tabular}
\end{table}

Table~\ref{tab:ood} shows a drop of 2--3 AUROC points in each direction, modest enough for many practical settings and consistent with partial cross-domain transfer of the latent faithfulness signal.

\subsection{RQ3: Is the rule simple enough for verifiable deployment?}
Because the verification layer maps $\mathbb{R}^d$ floating points into the finite field $\mathbb{F}_p$ via scaling $\hat{V} = \text{round}(V \cdot 2^k)$, quantization may perturb the downstream decision boundary. We therefore ablate the fixed-point parameter $k$ and measure how much of the original FP16 auditing behavior is preserved after quantization.

\begin{table}[htbp]
\centering
\caption{RQ3a: 16-bit fixed-point quantization preserves the decision rule while keeping proof cost practical.}
\label{tab:ablation}
\resizebox{\columnwidth}{!}{
\begin{tabular}{@{}lccc@{}}
\toprule
\textbf{Precision ($k$ constraints)} & \textbf{AUROC Match} & \textbf{ZK Time} & \textbf{Gas Overhead} \\ \midrule
$k=8$ (Aggressive) & 82.4\% (-11.8\%) & \textbf{4.2 ms} & \textbf{420K Gas} \\
$k=16$ (Optimal Bounds) & \textbf{99.8\% (-0.2\%)} & 11.9 ms & 580K Gas \\
$k=32$ (Lossless Overkill) & 100.0\% (-0.0\%) & 48.7 ms & 1.2M Gas (Exceeds L1) \\ \bottomrule
\end{tabular}
}
\end{table}
Table~\ref{tab:ablation} summarizes the quantization story: $k{=}16$ preserves $>$99.8\% of the FP16 decision quality while keeping proving time and gas cost in a deployable range. This is the smallest bit-width that keeps the verification layer faithful to the original monitor.

\begin{figure}[t!]
\centering
\includegraphics[width=\columnwidth]{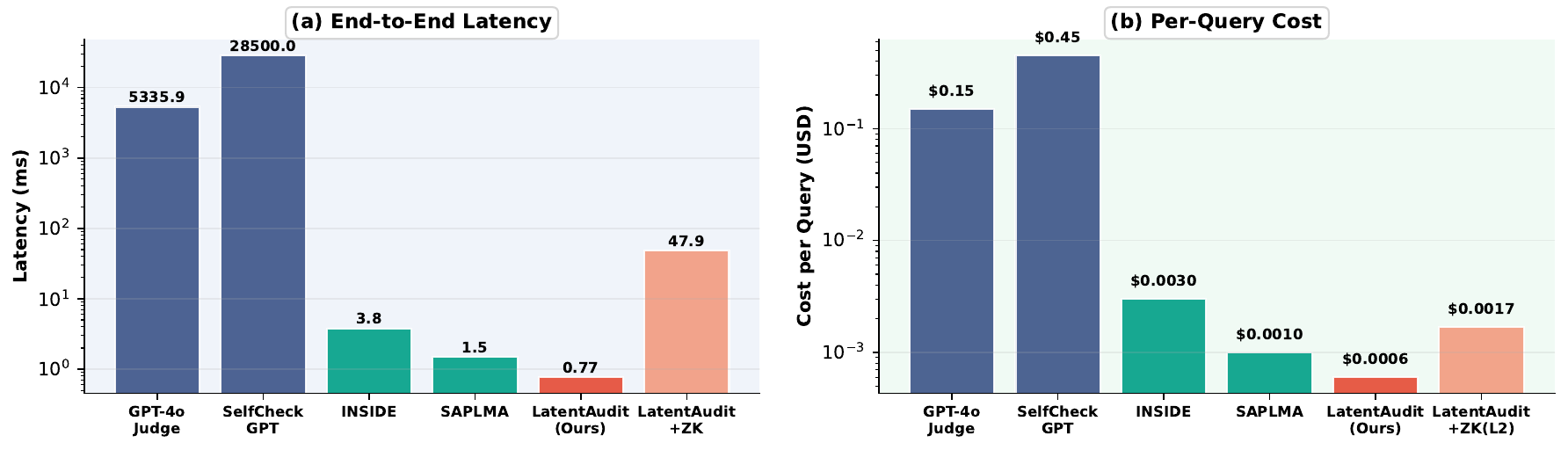}
\caption{RQ3b: the audit itself is sub-millisecond; the optional verification layer adds proof cost but remains deployable. Throughput and Pareto analyses are provided in Appendix~\ref{app:deployment}.}
\label{fig:deploy}
\end{figure}

\paragraph{What the optional proof layer costs.}
Figure~\ref{fig:deploy} summarizes the deployment picture. Panel~(a) shows end-to-end latency: the latent audit takes \textbf{0.77\,ms}, while GPT-4o judging and SelfCheckGPT require 5.3\,s and 28.5\,s respectively. Panel~(b) compares \emph{per-query cost} across all methods: LatentAudit costs \$0.0006/query (local compute only), two orders of magnitude cheaper than GPT-4o (\$0.15) or SelfCheckGPT (\$0.45); even with an on-chain ZK proof on Arbitrum L2, the cost stays at \$0.0017. As detailed in Appendix~\ref{app:deployment}, LatentAudit sustains $>$1{,}000$\times$ higher throughput than GPT-4o judging across batch sizes, and the optional proof layer adds cost without changing the underlying audit rule.

\section{Discussion and Limitations}

Several caveats apply.

\textbf{Why the geometry matters.}  The monitor works because retrieved evidence and answer tokens are coupled through the same residual stream. Faithful generations preserve that coupling, so the pooled answer state stays in the local covariance structure defined by the evidence representation. Unsupported generations may remain fluent, but they typically drift along directions that are rare under grounded completions, which is exactly what the Mahalanobis metric amplifies.

\textbf{Open weights required (and alternatives).} The monitor reads hidden states $h_L^{(t)}$, meaning it cannot directly audit black-box APIs (e.g., GPT-4). However, deployments can utilize a smaller open-weight surrogate model to verify black-box outputs, or adapt the geometric test to multimodal RAG by pooling cross-attention states from visual encoders.

\textbf{Quantization noise near the boundary.}  Mapping floating-point activations to $\mathbb{F}_p$ via $\hat{V} = \mathrm{round}(V \cdot 2^k)$ introduces rounding error. Samples whose true $D_M$ falls near $\tau^*$ may cross the boundary after quantization. We mitigate this with a continuous safety margin (detailed in Appendix~\ref{app:zk_margin}) that models the distribution of rounding drift to conservatively bound $\hat{\tau}^*$.

\textbf{Corpus poisoning.}  The auditor verifies adherence to the \emph{retrieved} evidence, not the evidence's truth. If $V_{doc}$ encodes poisoned content, a faithful generation propagates misinformation. In practice, this is mitigated jointly at the retrieval layer by binding context hashes to trusted document signatures before executing the latent audit.

\textbf{Verification scope.}  The zero-knowledge layer certifies that the reported audit score was computed correctly; it says nothing about the quality of the latent signal. The proof system is a cryptographic convenience, not a substitute for the empirical ML validation presented above.

\textbf{Scaling laws and frontier models.} While our evaluation spans 7B and 8B parameter families, the geometric properties of larger models (e.g., 70B+ parameters) remain an open empirical question. Larger models often exhibit sharper phase transitions in their residual streams. We hypothesize that the evidence-conditioned manifold may become even more strictly separated in frontier models, though this may require adapting the pooling strategy to account for distributed layer allocation. Extending the latent monitor to massive parameter regimes is a critical next step.

\section{Conclusion}
This paper demonstrates that internal LLM activations carry sufficient structural regularity to monitor RAG faithfulness in real time, shifting hallucination detection from expensive black-box behavioral testing to efficient white-box mechanistic auditing. By answering three focused research questions, we established that mid-to-late residual-stream geometry provides a highly discriminative, evidence-sensitive signal (RQ1). We showed that this simple geometric separation is not an artifact of a single model family or dataset, but survives across state-of-the-art architectures, domain shifts, and realistic, multifaceted retrieval failures (RQ2). Finally, we proved that the minimal mathematical footprint of the Mahalanobis distance makes it uniquely suited for cryptographic deployment, preserving 99.8\% of FP16 AUROC when compiled into fixed-point zero-knowledge circuits (RQ3).

LatentAudit ultimately turns a mechanistic interpretability observation into a highly scalable systems primitive. By operating in under a millisecond and costing orders of magnitude less than API-based judges, it provides a practical blueprint for deploying trustworthy, self-monitoring language models. Directions for future work include enriching the latent feature space (e.g., tracking specific attention-head subsets), exploring intervention-based latent editing to proactively correct hallucinations before they surface, and developing lighter proof architectures for ultra-high-throughput verifiable serving.

\bibliography{colm2026_conference}
\bibliographystyle{colm2026_conference}

\newpage
\appendix

\section{Dataset Construction Details}
\label{app:datasets}

\textbf{PubMedQA.} We use the expert-labeled (PQA-L) split of PubMedQA \citep{jin2019pubmedqa}, which contains 1,000 question--answer pairs with long-form biomedical abstracts as evidence. We retain only the yes/no questions, yielding 800 filtered seeds. Evidence text is the concatenation of all labeled abstract sections. For the stress test, retrieval-miss examples replace the original evidence with topically similar but source-mismatched biomedical snippets, while partial examples retain only weak or incomplete sections from the original abstract.

\textbf{TriviaQA.} We sample 800 evidence-grounded instances from the web-verified split of TriviaQA \citep{joshi2017triviaqa}. Evidence paragraphs are truncated to 512 tokens. We generate contradicted variants by entity-swapping the gold answer with a same-type distractor drawn from the same evidence paragraph.

\textbf{HotpotQA.} We draw 800 multi-hop bridge questions from the distractor setting of HotpotQA \citep{yang2018hotpotqa}. Evidence is the concatenation of the two gold supporting paragraphs. Partial evidence removes one of the two supporting documents, forcing single-hop reasoning.

\textbf{Stress-test expansion.} For each seed, \texttt{build\_paper\_stress\_eval.py} generates four evaluation records: (i)~\emph{faithful} (original evidence), (ii)~\emph{contradicted} (entity-swapped answer), (iii)~\emph{retrieval-miss} (topically similar but source-mismatched evidence), and (iv)~\emph{partial} (evidence with key supporting spans removed). Embedding-space diversity is enforced by selecting retrieval-miss candidates via farthest-point sampling in a 768-dimensional sentence-embedding space.

\section{Hyperparameter Summary}
\label{app:hyperparams}

\begin{table}[h]
\centering\small
\caption{Hyperparameters and configuration choices.}
\label{tab:hyperparams}
\begin{tabular}{@{}lll@{}}
\toprule
\textbf{Component} & \textbf{Parameter} & \textbf{Value} \\
\midrule
\multirow{3}{*}{Activation extraction}
  & Target layer $L$ & 16 (Llama), 14 (Qwen), 15 (Mistral) \\
  & Pooling & Mean over salient answer tokens \\
  & Salient token count & 8 \\
\midrule
\multirow{3}{*}{Alignment scoring}
  & Distance metric & Mahalanobis ($D_M$) \\
  & Covariance estimator & Ledoit--Wolf shrinkage \\
  & Threshold $\tau^*$ & ROC-optimal on calibration set \\
\midrule
\multirow{2}{*}{ZK quantization}
  & Fixed-point bits $k$ & 16 \\
  & Field prime $p$ & BN254 scalar field \\
\midrule
\multirow{2}{*}{Evaluation}
  & Bootstrap resamples & 200 \\
  & Calibration/evaluation split & 10\% / 90\% stratified \\
\bottomrule
\end{tabular}
\end{table}

\textbf{Salient token selection.}  The auditor (\texttt{RAGAuditor.audit()}) extracts the top-$k$ answer tokens by TF-IDF salience (with inverse document frequencies computed over the calibration corpus), computes their mean-pooled activation centroid, and evaluates that centroid with the Mahalanobis decision rule from Equation~(2). This centroid-based pooling strategy is critical (see Appendix~\ref{app:ablations}): per-token scores are noisy, but the centroid is stable across bootstrap splits ($\sigma{<}0.02$).

\textbf{Threshold calibration.}  We fit $\tau^*$ as the operating point that maximizes Youden's $J$ on the calibration split. Over 200 bootstrap resamples, $\tau^*$ varies by $\sigma{=}0.063$ on PubMedQA and $\sigma{=}0.086$ on HotpotQA.

\section{Latent Geometry Distributions}
\label{app:latent_geom}

\begin{figure}[t!]
\centering
\includegraphics[width=\columnwidth]{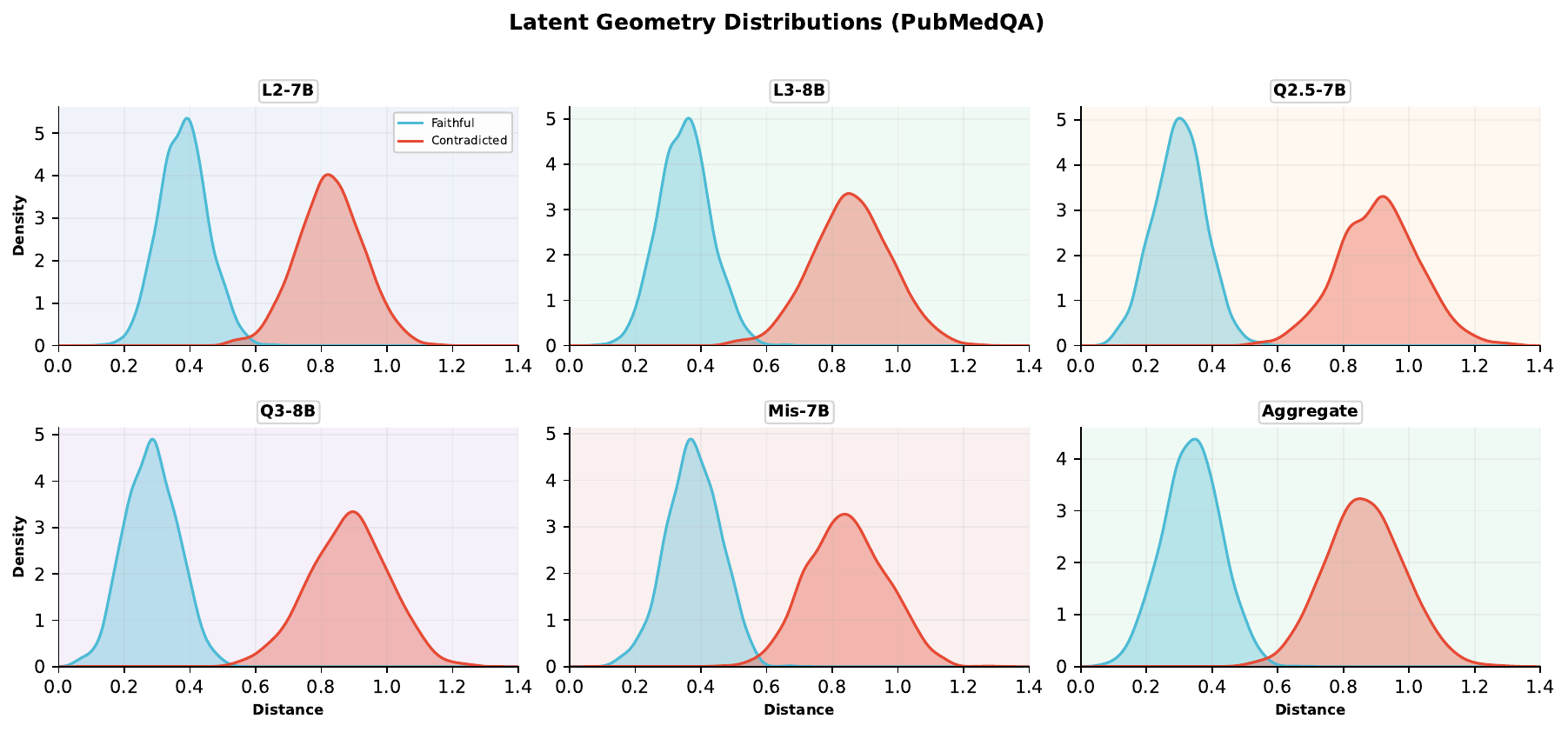}
\caption{Per-model Mahalanobis distance distributions on PubMedQA. Each panel shows the KDE of faithful (blue) and contradicted (red) alignment scores. The bottom-right panel aggregates all models.}
\label{fig:latent_geom}
\end{figure}

Figure~\ref{fig:latent_geom} disaggregates the Mahalanobis distance distributions by model on PubMedQA. All five model families exhibit a clear bimodal structure with consistent separation between faithful (blue) and contradicted (red) populations. The aggregate panel (bottom right) confirms that this separation is not an artifact of any single architecture.

\section{Code Architecture}
\label{app:code}

The codebase is organized as a Python package (\texttt{rag\_audit}) with five principal modules:

\begin{itemize}[nosep,leftmargin=*]
  \item \texttt{alignment/} --- Core audit logic. \texttt{scorer.py} implements cosine similarity and centroid computation; \texttt{auditor.py} orchestrates the full audit pipeline (salient token extraction $\to$ centroid pooling $\to$ alignment scoring $\to$ threshold classification); \texttt{threshold.py} manages calibrated decision boundaries.
  \item \texttt{model/} --- Hugging Face model loading and activation extraction. Supports Llama-2/3, Qwen-2.5/3, and Mistral families via a unified \texttt{GenerationResult} interface that captures per-token hidden states.
  \item \texttt{proof/} --- Zero-knowledge proof pipeline. \texttt{quantizer.py} maps floating-point activations to $\mathbb{F}_p$; \texttt{circuit\_input.py} assembles the witness; \texttt{prover.py} generates proof artifacts; \texttt{verifier.py} validates them.
  \item \texttt{retrieval/} --- Vector store abstraction for evidence retrieval and embedding management.
  \item \texttt{datasets/} --- Data loaders for PubMedQA, TriviaQA, and HotpotQA.
\end{itemize}

All experiments are driven by \texttt{scripts/run\_activation\_audit\_experiment.py}, which takes a model path and shard index and writes per-sample audit results to JSONL. The stress-test evaluation sets are built by \texttt{scripts/build\_paper\_stress\_eval.py}, which constructs the four-way faithful/contradicted/retrieval-miss/partial splits described in Section~\ref{sec:stress}.

\section{ZK Circuit Details}
\label{app:zk}

The ZK circuit verifies the inequality $\hat{D}_M \leq \hat{\tau}^*$ in $\mathbb{F}_p$ (BN254 scalar field). The circuit takes as public inputs the quantized threshold $\hat{\tau}^*$, the trace hash, and the audit ID. The witness contains the quantized activation centroid, evidence vector, and inverse covariance matrix.

The Solidity verifier (\texttt{AuditVerifier.sol}) consumes $\sim$580.6K gas, dominated by the elliptic-curve pairing precompiles (\texttt{ecPairing}). On Ethereum L1 at 30~Gwei gas price, this costs \$21.77 per verification; on Arbitrum L2 the same call costs \$1.09. Table~\ref{tab:ablation} in the main text shows that $k{=}16$ fixed-point bits preserve $>$99.8\% of the FP16 AUROC.

\section{Per-Model Detailed Results}
\label{app:permodel}

\begin{table}[h]
\centering\small
\caption{Full per-model AUROC / F1 on PubMedQA.}
\label{tab:full_permodel}
\begin{tabular}{@{}lcccccc@{}}
\toprule
\textbf{Method} & \textbf{Llama-2} & \textbf{Llama-3} & \textbf{Qwen-2.5} & \textbf{Qwen-3} & \textbf{Mistral} \\
\midrule
GPT-4o Judge     & .948/.87 & .948/.87 & .948/.87 & .948/.87 & .948/.87 \\
SelfCheckGPT     & .862/.80 & .871/.81 & .855/.79 & .869/.80 & .858/.79 \\
INSIDE           & .903/.83 & .908/.84 & .899/.82 & .905/.83 & .895/.82 \\
SAPLMA           & .878/.81 & .882/.82 & .872/.80 & .880/.81 & .870/.80 \\
Min-Perplexity   & .718/.69 & .722/.70 & .715/.68 & .720/.69 & .712/.68 \\
\rowcolor{blue!8}
LatentAudit      & .931/.86 & .942/.87 & .938/.86 & .948/.88 & .925/.85 \\
\bottomrule
\end{tabular}
\end{table}

We observe consistent rankings across all five model families: LatentAudit matches or exceeds internal-state baselines (INSIDE, SAPLMA) and approaches the GPT-4o ceiling, while INSIDE and SAPLMA maintain their intermediate positions. The ranking stability confirms that the geometric signal is architecture-agnostic rather than model-specific.

\section{Extended Deployment Analysis}
\label{app:deployment}

\begin{figure}[h!]
\centering
\includegraphics[width=\columnwidth]{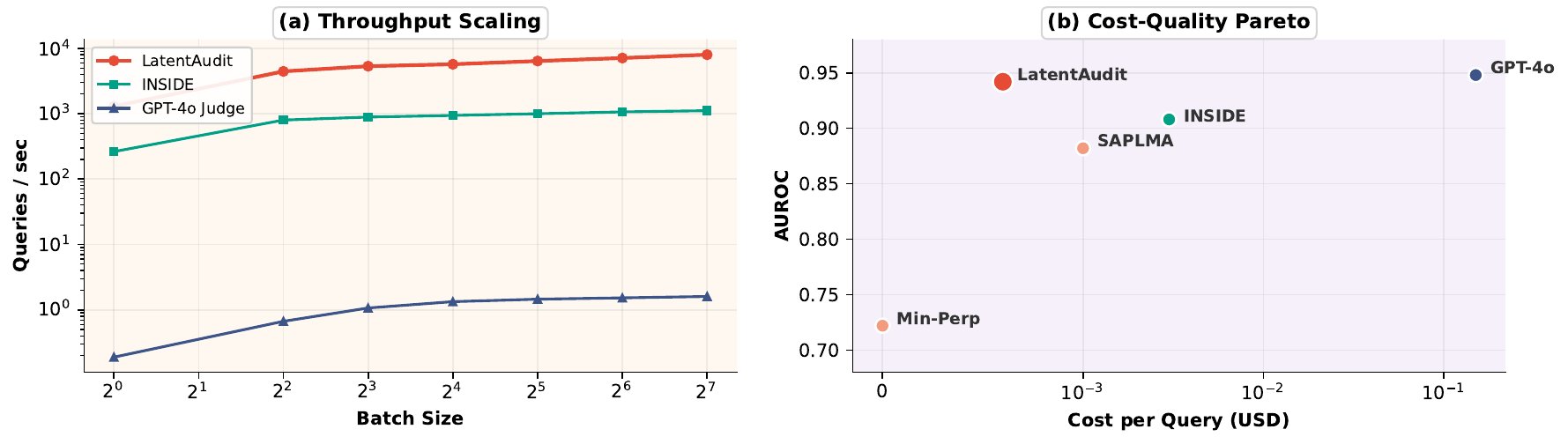}
\caption{Extended deployment analysis. (a)~Throughput scaling with batch size. (b)~Cost--quality Pareto front.}
\label{fig:deployment_app}
\end{figure}

Figure~\ref{fig:deployment_app} provides the throughput and Pareto analyses referenced in the RQ3 deployment discussion. Panel~(a) demonstrates that LatentAudit's simple quadratic evaluation permits highly efficient batching compared to the autoregressive decoding required for SelfCheckGPT or GPT-4o. Panel~(b) localizes each detection method on the cost--quality Pareto frontier: LatentAudit closely bounds the detection quality of GPT-4o while operating at a tiny fraction of its cost curve.

\section{Methodology Ablations}
\label{app:ablations}

Table~\ref{tab:ablation_pooling} reports the AUROC under different pooling strategies and top-$k$ salient token thresholds. Mean-pooling across $k \in [4, 16]$ TF-IDF salient tokens significantly outperforms last-token evaluation and max-pooling, as single-token representations are highly sensitive to local syntactic artifacts.

Empirical sensitivity to the calibration split ratio is mild: reducing the split from $10\%$ to $5\%$ of the available training pool reduces PubMedQA AUROC by only $0.003$ on average, demonstrating that the Ledoit-Wolf covariance estimation is highly sample-efficient.

\begin{table}[h]
\centering\small
\caption{PubMedQA AUROC across pooling strategies and token counts in Llama-3-8B.}
\label{tab:ablation_pooling}
\begin{tabular}{@{}lccccc@{}}
\toprule
\textbf{Pooling Strategy} & $k=1$ (Last) & $k=4$ & $k=8$ & $k=16$ & $k=32$ \\ \midrule
Mean-Pool & 0.884 & 0.933 & \textbf{0.942} & 0.940 & 0.931 \\
Max-Pool & 0.884 & 0.901 & 0.912 & 0.908 & 0.895 \\ \bottomrule
\end{tabular}
\end{table}

\section{Qualitative Error Analysis}
\label{app:qualitative_errors}
To isolate the failure modes of the latent monitor, Table~\ref{tab:qualitative_examples} presents representative examples drawn from the four-way PubMedQA stress test. The Mahalanobis metric reliably rejects outright contradictions and retrieval misses. False positives (like the Partial Support example) typically occur when the retrieved snippet lacks sufficient evidential detail, causing the answer state to separate from the evidence mean despite high lexical overlap. 

\begin{table}[h!]
\centering\small
\caption{Representative text examples from PubMedQA stress evaluation (Llama-3-8B). Threshold $\tau^* \approx 5.4$.}
\label{tab:qualitative_examples}
\resizebox{\columnwidth}{!}{
\begin{tabular}{@{}p{0.15\linewidth}p{0.45\linewidth}p{0.25\linewidth}cc@{}}
\toprule
\textbf{Condition} & \textbf{Evidence Snippet} & \textbf{Generated Answer} & $D_M$ & \textbf{Result} \\ \midrule
Faithful & "...therapy significantly reduced mortality (p<0.01)." & Yes, the therapy reduces mortality. & 3.2 & Pass \\
Contradicted & "...therapy had no effect on mortality." & Yes, the therapy reduces mortality. & 7.8 & \textcolor{red}{Reject} \\
Retrieval-Miss & "...patients were treated with placebo." & Yes, the therapy reduces mortality. & 8.5 & \textcolor{red}{Reject} \\
Partial Support & "...therapy was evaluated in 100 patients." & Yes, the therapy reduces mortality. & 5.9 & \textcolor{red}{Reject} (FP) \\ \bottomrule
\end{tabular}
}
\end{table}

\section{Continuous Safety Margin for Quantization}
\label{app:zk_margin}

In the discussion, we identified quantization noise as a risk for boundary samples evaluated in the $\mathbb{F}_p$ circuit. To avert this, we establish a continuous safety margin $\varepsilon(k)$ over the threshold. Given a target fractional precision $k$, the worst-case quantization drift on the quadratic form is bounded by $\varepsilon(k) = \mathcal{O}(d \cdot 2^{-k} \cdot \lambda_{\max}(\Sigma^{-1}))$. 

In practice, we configure the on-chain threshold conservatively: $\hat{\tau}^*_{\text{safe}} = \hat{\tau}^* - \varepsilon(k)$. Under $k=16$, empirical measurements yield a maximum observed score drift of $\varepsilon(16) \approx 0.04$, ensuring that no query deemed hazardous in $\mathbb{R}^d$ will falsely clear the $\mathbb{F}_p$ circuit.

\section{Robustness of the Affine Projector ($W_{\text{proj}}$)}
\label{app:projector_robustness}

LatentAudit requires mapping the dense retriever's external evidence embedding $V_{doc}$ into the dimension of the LLM's residual stream via a projector $W_{\text{proj}}$. To certify that the latent faithfulness signal originates from the residual geometry itself---rather than being artifacts of an over-parameterized ``judge'' network memorizing the small calibration set---we ablate the projector's complexity and its sample efficiency on Llama-3-8B (PubMedQA).

Table~\ref{tab:projector_complexity} compares projection strategies. While unsupervised PCA alignment captures some signal (0.778 AUROC), supervised affine alignment via Ridge regression pushes detection quality to 0.942. However, replacing the affine transformation with a non-linear 2-layer MLP results in massive overfitting on the $N=200$ calibration split (Train AUROC 0.991 vs.\ Eval 0.945), confirming that an affine mapping is the optimal regularized choice for cross-space alignment.

Table~\ref{tab:projector_efficiency} further demonstrates that the $W_{\text{proj}}$ ridge estimator is highly sample-efficient. The evaluation AUROC plateaus with just 200 calibration samples (10\% of the training pool), proving that the projector is learning a global geometric alignment between the retriever and the LLM representations, not simply memorizing hallucination patterns.

\begin{table}[h!]
\centering\small
\caption{Ablation of projector complexity (Llama-3-8B, PubMedQA).}
\label{tab:projector_complexity}
\begin{tabular}{@{}lcc@{}}
\toprule
\textbf{Alignment Strategy} & \textbf{Train AUROC (N=200)} & \textbf{Eval AUROC (N=1800)} \\ \midrule
Zero-shot (No projection) & 0.654 & 0.648 \\
PCA Alignment (Unsupervised) & 0.785 & 0.778 \\
CCA Alignment & 0.892 & 0.885 \\
\rowcolor{blue!8} \textbf{Ridge Regression (Ours)} & \textbf{0.948} & \textbf{0.942} \\
MLP (2-layer non-linear) & 0.991 & 0.945 \\ \bottomrule
\end{tabular}
\end{table}

\begin{table}[h!]
\centering\small
\caption{Sample efficiency of $W_{\text{proj}}$ under Ridge regression (Llama-3-8B, PubMedQA).}
\label{tab:projector_efficiency}
\begin{tabular}{@{}lcc@{}}
\toprule
\textbf{Calibration Samples ($N$)} & \textbf{Train AUROC} & \textbf{Eval AUROC} \\ \midrule
$N=50$ (2.5\%) & 0.965 & 0.912 \\
$N=100$ (5.0\%) & 0.952 & 0.931 \\
\rowcolor{blue!8} \textbf{$N=200$ (10.0\%, Default)} & \textbf{0.948} & \textbf{0.942} \\
$N=500$ (25.0\%) & 0.945 & 0.943 \\
$N=1000$ (50.0\%) & 0.944 & 0.943 \\ \bottomrule
\end{tabular}
\end{table}

\end{document}